\documentclass[final]{siamonline220329}
\usepackage{amsmath,amsfonts,dsfont,bbm}
\usepackage{listings}
\usepackage[justification=justified,font=footnotesize]{caption}
\usepackage{hyperref,cleveref,xcolor}
\usepackage{graphicx}
\usepackage{subcaption}
\usepackage{booktabs}

\def\R{\ensuremath{\mathbb{R}}}

\def\Rn{\ensuremath{\mathbb{R}^n}}
\def\Rm{\ensuremath{\mathbb{R}^m}}

\newcommand{\bderv}[1]{B-Deriv#1}  
\newcommand{\mujoco}[1]{MuJoCo#1}
\newcommand{\pybullet}[1]{PyBullet#1}
\newcommand{\St}{\mathfrak{X}}
\newcommand{\Cont}{\mathcal{C}}
\newcommand{\Hdom}{\mathcal{H}}
\newcommand{\Fl}{\mathcal{F}}
\newcommand{\Gua}{\mathfrak{g}}
\newcommand{\Nei}[1]{N^{#1}}
\newcommand{\Neps}{\Nei{\varepsilon}}

\begin{document}

	\maxdepth=0pt

	\author{Marion Anderson\thanks{Department of Electrical and Computer Engineering, %
		University of Michigan, Ann Arbor, MI, USA%
	  (\email{marand@umich.edu})}
	\and Shai Revzen\thanks{Department of Electrical and Computer Engineering, %
		University of Michigan, Ann Arbor, MI, USA%
	  (\email{shrevzen@umich.edu})}
}

	\title{\LARGE \bf Rapid Integrator for a Class of Multi-Contact Systems
        \thanks{\funding{This work was funded by the D. 
Dan and Betty Kahn Foundation Autonomous Systems Megaproject, NSF CPS 2038432, and ARO MURI W911NF-17-1-0306.}}}

	\maketitle

	\begin{abstract}
    Many problems in robotics involve creating or breaking multiple contacts nearly simultaneously or in an indeterminate order. 
We present a novel general purpose numerical integrator based on the theory of Event Selected Systems (ESS). 
Many multicontact models are ESS, which has recently been shown to imply that despite a discontinuous vector field, the flow of these systems is continuous, piecewise smooth, and has a well defined orbital derivative for all trajectories, which can be rapidly computed. 
We provide an elementary proof that our integrator is first-order accurate and verify numerically that it is in fact second-order accurate as its construction anticipated. 
We also compare our integrator, implemented in NumPy, to a MuJoCo simulation on models with 2 to 100 contacts, and confirm that the increase in simulation time per contact is nearly identical. 
The results suggest that this novel integrator can be invaluable for modelling and control in many robotics applications. 
\end{abstract}

	\section{Introduction}
A defining property of robots is the ability to physically interact with their environment. 
To control these interactions, roboticists usually rely on models in the form of equations of motion, which allow a computer to anticipate the trajectory robots and objects will follow. 
When this trajectory's first-order dependence on parameters and initial conditions (its ``orbital derivative'') is computable, the models become much more amenable to planning and control optimization \cite{posa-2014-rigid-contact-optim,payne2020uaskf,goebel2009hybridoverview,wendel-2012-hybrid}

These equations of motion are typically represented using ordinary differential equations (ODEs). 
The ODEs which describe robot tasks where the robot makes and breaks contact with its environment, such as legged locomotion or dexterous gripping, commonly use ``hard contacts'' to describe the process of making or breaking contact \cite{posa-2014-rigid-contact-optim}. 
These models produce ODEs with a jump discontinuity at the moment of contact (a ``hybrid transition'' at the crossing of a ``guard'' surface) and are often viewed as a type of ``hybrid systems'' or ``non-smooth systems'', which require special attention to integrate accurately (see \cite{goebel2009hybridoverview} for terminology). 
One approach for dealing with this complication has been to replace hard contact models with so-called ``soft contact'' approximations \cite{li-2001-soft-contact-review,mujoco,Raff2023HardVS}, whose smoothness allows first-order dependencies to be numerically computed. 
Unfortunately, whether hard or soft, it is common for several contacts to be made or broken nearly simultaneously. 
Examples include the tripod walking gaits of RHex \cite{altendorfer-2001-rhex}, iSprawl \cite{sangbae-2005-isprawl}, and Dante II \cite{barres-1999-dante2}, humanoid grasping such as the Utah/MIT Dextrous Hand \cite{jacobsen-1984-utah/mit-hand} and the Karlsruhe dextrous hand \cite{wohlke-1994-karlsruhe}, and the multitude of three-fingered dextrous manipulators \cite{bicchi-2000-dextrous-difficult,jongkind-1993-dext,okada-1979-dext}. 

Although there have been numerous results recovering first-order approximations for trajectories undergoing a single isolated hybrid transition \cite{filippov-1988-discontinuous, ivanov-1998-stability}, or even a pair of near simultaneous hybrid transitions \cite{ ivanov-1998-stability}, first order approximations for trajectories undergoing three or more near-simultaneous transitions have only recently been published \cite{Burden-2015-multi,council-2021-representing}. 
The complexity of the problem with three or more transitions stems from the combinatorial growth in the ordering of transition events \cite{hybridoverview}. 
For even the simplest cases, it can be shown that the entire factorial number of transition orderings is reachable by infinitesimally perturbing an initial condition. 
Recent work has shown that common near-simultaneous collisions which arise in robotics, under mild technical conditions, admit a first-order approximation \cite{Burden-2015-multi} called the Bouligand Derivative (herein referred to as the ``\bderv{}''). 
For these systems, the \bderv{} is continuous, piecewise linear, and computationally equivalent to the one-sided difference quotient limit (``directional derivative'') \cite{Burden-2015-multi}. 
\cite{council-2021-representing} developed an efficient algorithm for computing the \bderv{} from this result. 
This algorithm does not require the root finding numerical machinery commonly found in contact solvers \cite{council-2021-representing,dart,pybullet,odesim,mujoco}. 

\begin{figure}[h]

\centering

\includegraphics[width=0.8\columnwidth]{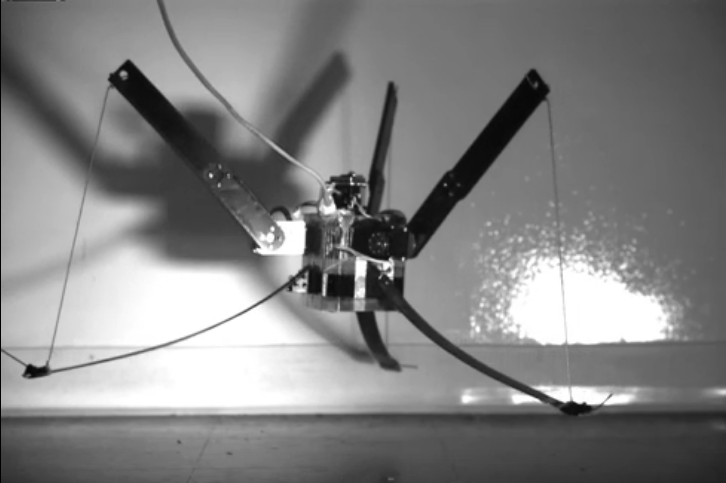}

\caption{A three-legged hopping robot we developed, which motivated much of this work \cite{mclaughlin-2020-hopping,anderson-2020-dw}. 
It ``preloads'' its elastic legs prior to ground contact using a cable assembly. 
The cable assembly can only exert force upward on the legs. 
The legs immediately exert a force on the ground at the moment of contact. 
As a result, the hopper contact force model is affine, and cannot be readily solved using the machinery for linear complementarity problems \cite{brogliato2003perspectives}.}

\label{fig:hopper}

\end{figure}

Root finding is known to be a significantly expensive computation in solving the mechanical forces that arise in multi-contact, and has drawn much attention from the robotics community \cite{brogliato2003perspectives,bicchi-2000-dextrous-difficult}. 
Despite its slow speed, root finding is used to achieve sufficient accuracy for simulation and control. 
The speed-accuracy tradeoff is a common concept in numerical methods, and is often used when evaluating numerical methods against one another \cite{erez2015simcompare}. 
Removing root finding machinery without sacrificing (too much) accuracy could greatly reduce some of the need for computing power in robotics, enabling faster simulation and real-time control with slower (cheaper) on-board computers. 

\subsection{Related Work}
Regardless of the contact model used, hybrid systems describing contact events have historically been treated by roboticists in three ways:
with impact maps (sometimes called ``reset maps'') \cite{lygeros2001hybridautomata,westervelt2003hzd}, as linear complimentarity problems (LCPs) \cite{anitescu-1997-lcp-coloumbcontact,fazeli-2017-contact-friction}, and as multi-valued maps, often called ``differential inclusions'' \cite{Moreau1988DI,posa-2016-diffincl}. 
As each of these formalisms seeks to model similar physical phenomena, there has been some work to find equivalence relations between them \cite{brogliato2006equivalence}. 

LCPs are a broad class of matrix optimization problems, with much work done by Cottle starting in the late 1960s \cite{cottle-1979-lcp,cottle1970lcp,cottle2009lcp}. 
As such, LCPs benefit from a multitude of efficient solvers \cite[Chps. 
4,5]{cottle2009lcp}. 
They have much attention in robotics due to their natural extension to resolving the forces that occur in multi-contact \cite{anitescu-1997-lcp-coloumbcontact,fazeli-2017-contact-friction,lotstdet-1982-mechconstraints}. 
Although LCPs have such a natural extension to sticking contact, they can only approximate the forces for sliding contacts with Coloumb friction \cite{anitescu-1997-lcp-coloumbcontact}, and cannot represent nonlinear contact dynamics \cite{brogliato2003perspectives}. 
Regardless, they have achieved prolific use in both research and industrial settings \cite{dart,pybullet,odesim}. 
Differential inclusions, first introduced to engineering by Moreau \cite{Moreau1988DI} can treat friction directly, and are considered to most accurately describe the hybrid phenomena arising in robotics \cite{Kiseleva2018DIMech,posa-2016-diffincl,brogliato2003perspectives}. 
However, differential inclusions have not received much attention for robot simulation. 
Modern rigid body software simulation suites either use soft contact models (\mujoco{} \cite{mujoco}), LCP solvers (DartSim \cite{dart}, \pybullet{} \cite{pybullet}), or a combination of both (Open Dynamics Engine \cite{odesim}). 
Impact maps apply discontinuous (impulsive) updates to the vector field, and have primarily been used to model the single-contact phenomena that occur in bipedal locomotion \cite{posa-2016-diffincl,westervelt2003hzd,gregg20093dcompass}. 
Recent work \cite{Burden-2015-multi,council-2021-representing} suggests that modelling multi-contact using impacts maps may be more tractable than previously thought. 
Impact maps are the most similar formalism to the work presented in this manuscript. 

\subsection{Contributions}

We present, to our knowledge, the first integration algorithm to combine conventional ODE integration within a single ``hybrid domain'' with rapid resolution of multi-contact events that (at the limit) reproduces the correct orbital derivative. 
We describe its practical use in Sec. \ref{sec:usage} and its implementation alongside an example code in Sec. \ref{sec:algorithm}. 
We numerically assess its accuracy in Sec. \ref{sec:order}, compare its performance to the robotic simulation software \mujoco{} in Sec. \ref{sec:mujoco}, and provide a simple proof of correctness to first-order in Sec. \ref{sec:proof}. 

\section{Integrator Usage}\label{sec:usage}

This section is intended as a ``manual'' for the practitioner who wishes to use our integrator with their hybrid system and to provide a precise definition of requirements. 
See Sec. \ref{sec:mujoco} for modelling examples. 

\subsection{Mathematical Preliminaries}

Let $\St \subseteq \Rn$ be a compact domain. 
Let $\Hdom \subseteq \Rm$ be an open neighborhood of $0 \in \Rm$. 
\begin{align}
    f &: \St \times \Hdom \to \Rn, \tag*{(hybrid) dynamics vector field}\\
    h &: \St \to \Hdom, \tag*{event function(s)}
\end{align}
As $f$ and $h$ fully define the dynamics of interest, we will refer to the pair $(f,h)$ as the ``system.''
We define the ``guard'' $\Gua_k$ to be the zero-level set of $h_k(\cdot)$, the $k$-th component of $h$, i.e. $\Gua_k := \{ x \in \St \;|\; h_k(x) = 0 \}$. 
We will consider various (compact) ``$\varepsilon$-neighborhoods'' of $\Gua_k$ defined using $h_k(\cdot)$ by $\Neps := \{x \in \St \;|\; -\varepsilon \leq h(x) \leq \varepsilon\}$ for $\varepsilon\geq0$. 

We further require that:
\begin{enumerate}
    \item $f(x,h)$ is $\Cont^p$ smooth in its first parameter. 
    \item $f(x,y) = f(x,y')$ if $y_k y'_k > 0$ for all $k$. 
    In other words, the $y \in \Hdom$ dependence of $f(\cdot,y)$ is only on the signs of the components of $y$. 
    \item $h(\cdot)$ is $\Cont^p$. 
    \item There exists some $c>0$ such that for all $y \in \Hdom$, $k$, and for all $x\in\Nei{\nu_k}$ we have $\nabla h_k(x) \cdot f(x,y) > c$.\footnote{This is known as the ``liveness condition'' of ESS \cite[Defn. 
1]{Burden-2015-multi}.}
    \item For every $k$ we require a $\nu_k>0$ such that 
    for all (fixed) $b \in \Hdom$, and $x_0 \in \Nei{\nu_k}$, the unique solution for the ODE $\dot x = f(x,b)$ that passes through $x_0$ exists for the interval $[-h_k(x_0)/c, h_k(x_0)/c]$. 
    This condition ensures that all trajectories starting in $\Nei{\nu_k}$ cross $\Gua_k$. 
\end{enumerate}

By abuse of notation we define $f(x) := f(x,h(x))$, and note that $f(x)$ is smooth except perhaps at the guards $\Gua_k$. 
Since those are proper co-dimension 1 manifolds (liveness ensures $\nabla h_k \neq 0$), the dynamical system $\dot x = f(x)$ is a hybrid dynamical system in the sense of \cite{wendel-2012-hybrid} with guards $\Gua_k$. 

\subsection{Practical Use}

Our integrator requires definitions of $f$, $h$, the Jacobian of $h$, $Dh : \St \to L(\Rm,\Rn)$, and a precision parameter, $\varepsilon>0$.  
Note that the Jacobian need not be closed-form. 

After choosing appropriate event functions, $\varepsilon$ provides the primary means to trade off speed and accuracy. 
Increasing $\varepsilon$ reduces the number of integration steps performed at the risk of greater integration error and vice-versa. 
For any state $x_p\in\Nei{\nu_\varepsilon}$, the user should work to ensure $\|Dh(x_p) \cdot f(x_p)\|_\infty\leq\varepsilon$. 
This is always achievable because given some $\bar h$, there exists $\mu_k:=\sup_\St \nabla{\bar h}_k(x) \cdot f(x) < \infty$ because $\St$ is compact. 
Define $h_k:=\varepsilon/\mu_k {\bar h}_k$ to obtain the result. 

\section{Algorithm}\label{sec:algorithm}

\subsection{Implementation}\label{subsec:implementation}
To our knowledge, the integration algorithm presented in this section is the first to combine conventional ODE integration within a single hybrid domain with rapid resolution of multi-contact events using the \bderv{}. 
In the case of ESS, the \bderv{} is the first-order approximation to the flow, even in those locations where it is not classically (Fr\'echet) differentiable \cite{Burden-2015-multi}. 
We use the \bderv{} to resolve multi-contact events using tangent line approximations, effectively ``projecting'' the current state through the guards. 
Our algorithm performs conventional integration and \bderv{} integration in an alternating fashion. 
As such, we will refer to conventional integration as the ``integration step'' and \bderv{} integration as the ``projection step.'' 
Due to this alternating behavior, our algorithm utilizes a loop structure in software, shown in Listing 1. 
At the start of each loop, the trajectory has been integrated to time $t_q$ and state $x_q$. 

\begin{figure}[h!]

\centering

\includegraphics[width=0.8\columnwidth]{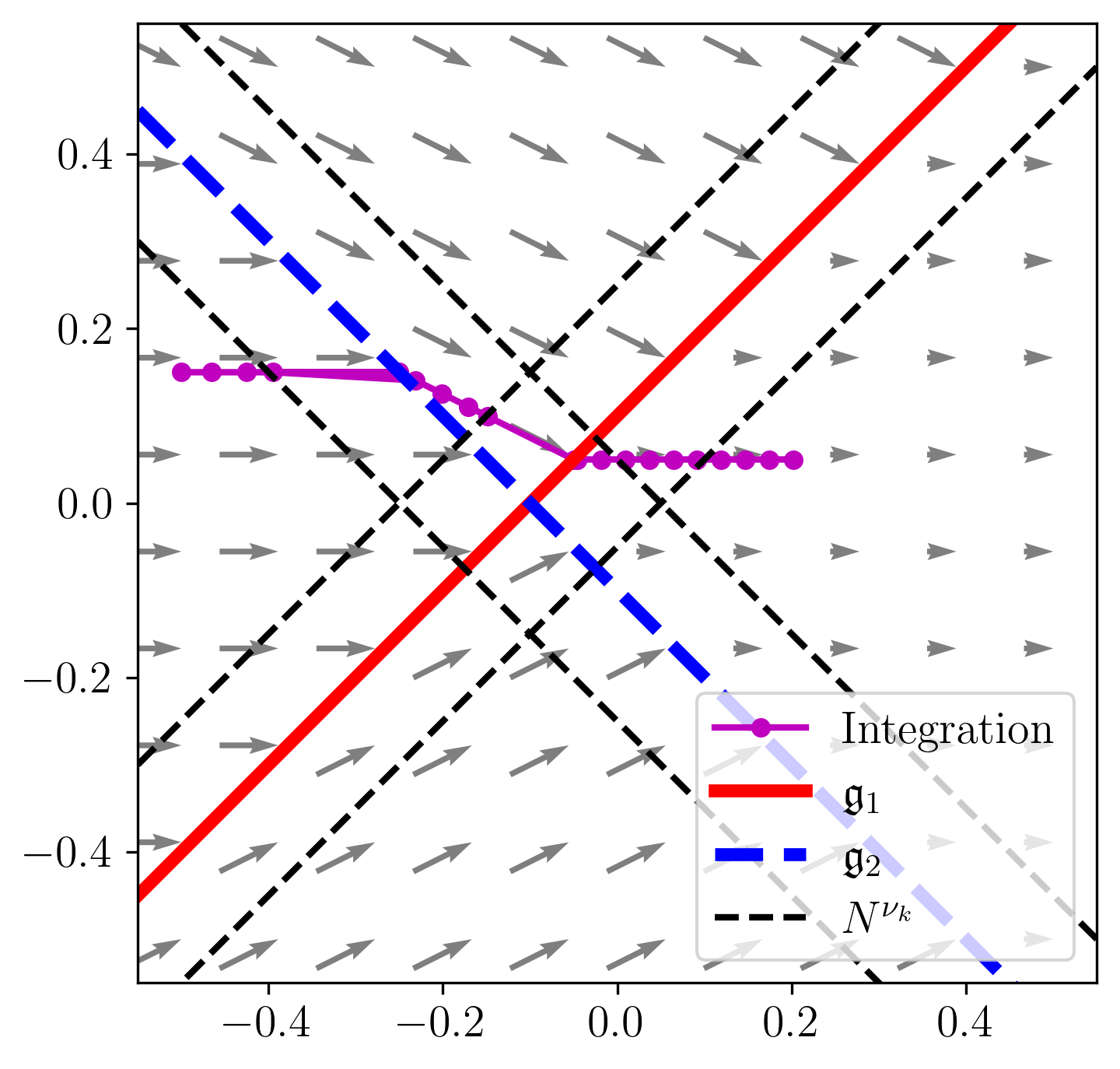}

\caption{Our algorithm integrating a piecewise-constant ESS (purple dotted line). 
The system has two guards, $\Gua_1$ (thick red line) and $\Gua_2$ (dashed thick blue line) with non-zero slope and intersect at a point away from the origin. 
The integrator projects (purple line) when the trajectory enters $\Nei{\nu_k}$, otherwise it performs conventional integration (purple dots). 
}

\label{fig:projection}

\end{figure}

By requirement (4) when hybrid transitions occur they are (locally) irrevocable.  This implies that we need only consider each guard once, at most. 
Our algorithm therefore maintains a pool of candidate guards which have not been crossed. 
In the context of the projection step, we will refer to candidate guards simply as guards for brevity. 
If $x_q$ lies in $\Nei{\nu_k}$ of some guards, we determine (to first order) which guard will be crossed first, and project to that guard with the first order approximation of the impact map as in \cite{westervelt2003hzd}, removing it from the candidate pool. 
When $x_q$ lies in $\Nei{\nu_k}$ for multiple guards, the loop structure of our algorithm naturally performs the projection step multiple times consecutively, equivalent to projecting across the guards in one computation. 
The projection operation is the exact impact map for a constant vector field dynamic crossing a co-dimension 1 hyper-plane guard. \cite{council-2021-representing} showed that a suitably chosen piecewise-constant vector field (the ``corner vector field'') has the same orbital \bderv{} as the original ESS did, and our algorithm employs a similar intuition. 
If $x_q$ does not lie in any $\Nei{\nu_k}$ we integrate forward with a conventional ODE integrator until the trajectory enters any $\Nei{\nu_k}$ (we used dopri5 \cite[II.4, p.178]{hairer1993solving}; most other integration schemes could also be used). 
To detect when the trajectory enters some $\Nei{\nu_k}$, the conventional integrator must provide a means to detect such an event and terminate integration. 
We used a termination condition detecting if the trajectory had entered $\Nei{\nu_k}$ of at least one guard. 

In common rhythmic hybrid systems, such as a 1D hopping robot, the same guard (the ground) can be crossed multiple times in alternating directions (stance-flight and flight-stance transitions) over a single trajectory. 
To integrate such a system  through multiple hops (guard crossings), our algorithm needs to re-add guards to the candidate pool under certain conditions. 
In our current implementation, when the candidate pool is empty, and the integration step has just completed, we re-add any guards with negative-valued event functions. 
In future implementations, we would like to generalize this behavior to a ``guard reset matrix,'' which would allow us to specify how crossing one guard re-enables other guards to become candidates. 
For the 1D hopper, crossing the flight-stance guard indicates the trajectory may soon cross the stance-flight guard and that it should be reset. 
\vspace*{-25pt}
\begin{tabular}[h]{c}
\hline
\vspace{2em}
\begin{lstlisting}[caption=Python implementation of our algorithm., captionpos=b]
import numpy as np
import favorite_integrator as dopri5

def integrate(f, h, Dh, eps, x0, t0, tf):
    # Termination Condition
    def ntr_Nvk(x1, x2):
        return ((h(x1) < -eps) & (h(x2) >= -eps)).any()
    dopri5.term_early_func = ntr_Nvk

    # Begin Integration
    x0 = np.asarray(y0)
    tq, xq = t0, x0.copy()
    tlog, xlog = [tq], [xq.copy()]
    crossmsk = h(x0) >= 0
    while tq < tf or not all(crossmsk):
        hx = h(xq)
        if not any((hx > -eps)[~crossmsk]):
            tq, xq = dopri5(xq, tq, tf)
            tq, xq = tq[1:], xq[1:]
            if all(crossmsk):
                crossmsk = h(xq[-1]) >= 0
        else:  # Projection Step
            G, e = f(xq, hx), Dh(xq)
            eG = np.dot(e, G)
            dt = -hx / eG
            dt[crossmsk] = inf
            j = argmin(dt)
            dt = dt[j]
            dx = G * dt
            tq = [tq + dt]
            xq = [xq + dx]
            crossmsk[j] = True
        # End Loop / Next Iteration
        tlog.append(tq)
        xlog.append(xq)
        tq, xq = tq[-1], xq[-1]
    return np.hstack(tlog), np.vstack(xlog)
\end{lstlisting}
\label{lst:code}
\end{tabular}

\subsection{Differences from Council et al \cite{council-2021-representing}}\label{subsec:gcouncil}

We derived our projection step from the algorithm presented by Council et al in \cite[Fig. 2.1]{council-2021-representing}. Our projection step differs in where we sample the vector field and event functions, and the assumptions we make about the trajectory's path through the event functions. 
A collection of $m$ event functions $\{h_k\}$ for $k\in\{1,\dots,m\}$ which all take the value $0$ at a point $\rho$ are called an ``($m$-dimensional) ESS complex (crossing at $\rho$).''
Council's algorithm assumes that the system has only one ESS complex, that the \bderv{} and $Dh$ at $\rho$ are already known, and that the coordinates have been transformed such that $\rho$ is the origin \cite{council-2021-representing}. 
We sought to relax these restrictions to support systems with multiple ESS complexes and to minimize the additional modelling requirements beyond defining a system's event functions. 
To that end, we do not assume knowledge of $\rho$ and instead sample $h$, $Dh$, and $f$ at a point in $\Nei{\nu_k}$ lying along the trajectory. 

\section{Order Test}\label{sec:order}

Prior to proving properties of a numerical method, it is often useful to evaluate the method on a toy problem with a known exact solution. 
Such an evaluation is doubly useful both to build intuition for the method's behavior and to verify that the implementation lacks obvious errors. 
Numerical integrators are commonly evaluated on the relationship between their step size and normed integration error as bounded by an integer-power polynomial \cite[Chp II]{hairer1993solving}. 
The degree of this polynomial minus one is called the integrator's ``order'' \cite[Chp II]{hairer1993solving}. 
Given an exact solution $y$, initial condition $y_0$, initial time $t_0$, and step size $\delta t$, an integrator approximates approximates the exact value $y(t_0+\delta t)$ as $y_1$. 
For an integrator of order $p$ and some constant $K$, the error bound takes the form
\[||y(t_0+\delta t) -y_1||\le K\delta t^{p+1} \] \cite[Eq 1.10, p.134]{hairer1993solving}. 

\begin{figure}[h]

\centering

\includegraphics[width=0.9\columnwidth]{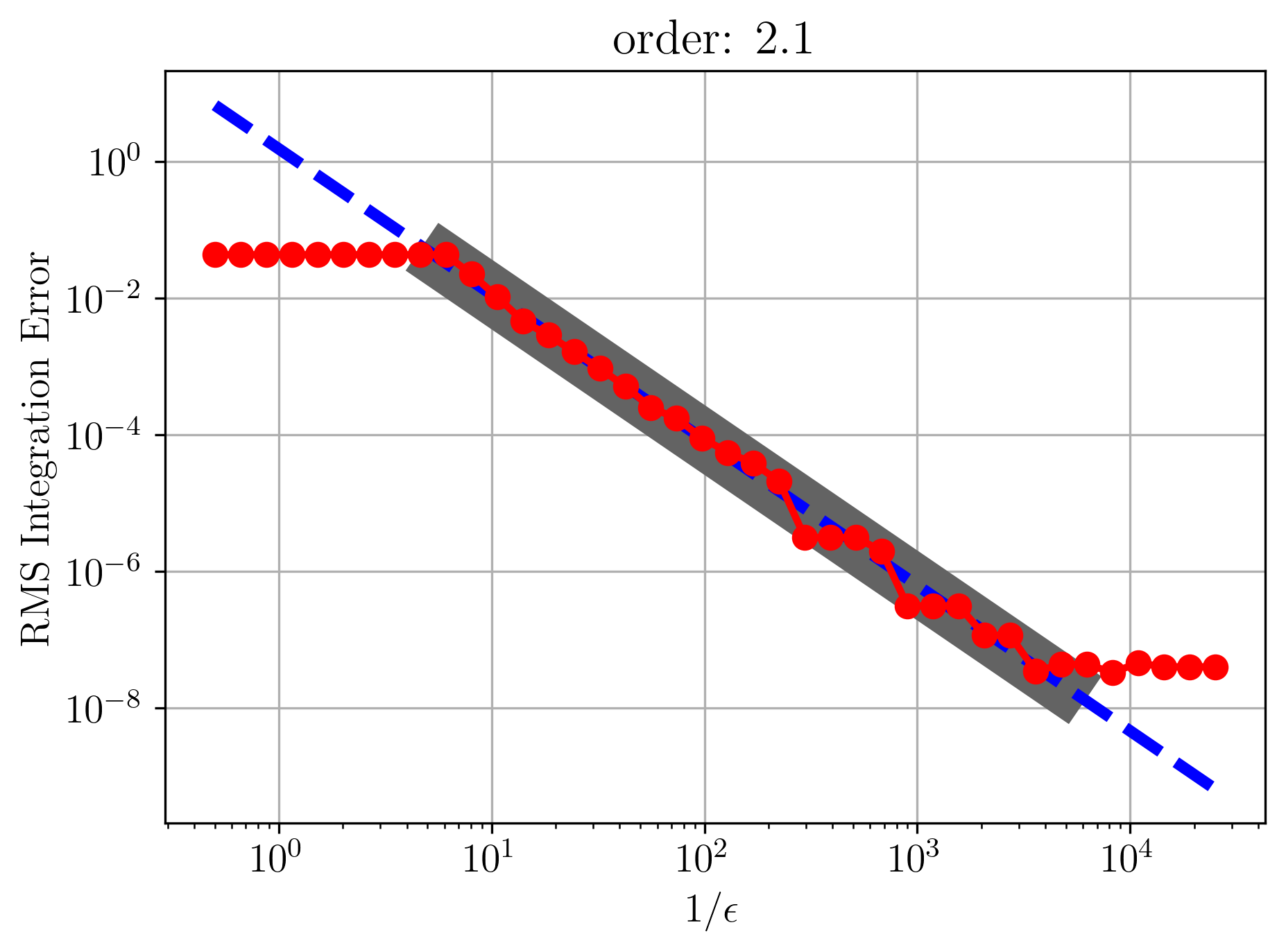}

\caption{Log-log plot of our algorithm's RMS integration error versus $1/\varepsilon$ (dotted red line) for an initial value problem. 
The data interval used to determine the order (gray rectangle) and the fit extended over the entire interval considered (dashed blue line) are overlaid. 
The integrator achieved second-order accuracy, indicated by the slope of the blue line. 
The flat tails for small and large $1/\varepsilon$ are expected. 
}

\label{fig:order}

\end{figure}

We sought to empirically assess the order of our algorithm's projection step as a function of $\varepsilon$. 
Our integration step utilized the 5th-order dopri5 integrator \cite[II.4, p.178]{hairer1993solving}, and the related \bderv{} projection algorithm developed by Council et al has been proven to be 2nd-order in $\varepsilon$ \cite{council-2021-representing}. 
As such, we expected our projection step to be at most 2nd-order. 
From this, we expected that any trajectory integration error should be dominated by the error of our projection step, and not by our integration step. 
For such an assessment, we needed a differential equation of sufficient complexity which still had a known closed-form flow. 
Linear and affine differential equations possess readily-computable exact solutions through the matrix exponential, making them a natural choice. 
For simplicity, we chose our guards to be the coordinate planes. 
We constructed a 3D system of affine ordinary differential equations, described in Tab. \ref{tab:order}

\begin{table*}[h!]
    \centering
    \begin{subtable}[t]{0.35\columnwidth}
    \centering
    \begin{tabular}[t]{c|cc}
        & $x<0$ & $x\ge0$ \\
    \midrule                                                  
        $y<0$ & $-y+1$ & $-2y+1$ \\
        $y\ge0$ & $y+1$ & $10x+1$ \\
    \bottomrule
    \end{tabular}
    \caption{$\dot{x}$}
    \label{tab:table1_a}
\end{subtable}%
\begin{subtable}[t]{0.25\columnwidth}
    \centering
    \begin{tabular}[t]{cc}
        $x<0$ & $x\ge0$ \\
    \midrule                                                  
        $x+1$ & $x/2+2$ \\
        $-x+1$ & $y+1$ \\
    \bottomrule
    \end{tabular}
    \caption{$\dot{y}$}
    \label{tab:table1_b}
\end{subtable}%
\begin{subtable}[t]{0.25\columnwidth}
    \centering
    \begin{tabular}[t]{cc}
     $z\le0$ & $z>0$ \\
    \midrule                                                  
    $3z-1$ & $-z-1$  \\
    $3z-1$ & $-z-1$  \\
    \bottomrule
    \end{tabular}
    \caption{$\dot{z}$}
    \label{tab:table1_c}
\end{subtable}%
\caption{\footnotesize The 3D system of affine ODEs used for our order test. 
The $x$ velocity (a), $y$ velocity (b), and $z$ velocity (c) are grouped separately.}
\label{tab:order}
\end{table*}

We integrated the trajectory with initial condition $(x,y,z)=(-0.4,-0.15,0.3)$. 
This trajectory crossed all guards during the interval $t\in(0,0.5)$. 
We computed integration error as the root-mean squared error (RMS) of the integrated trajectory from the exact trajectory. 
We found the RMS to be bounded by a polynomial of order at least $2.1$. 
From this experiment, we expect our algorithm to be 2nd-order, as $\mathrm{RMS}(\varepsilon)\le K\varepsilon^{2.1}<K\varepsilon^3$, see Fig. \ref{fig:order}. 

\section{Performance Comparison with \mujoco{}}\label{sec:mujoco}

\subsection{Calibration}

To assess the practical merit of our algorithm, we sought to compare it with a state-of-the-art simulation framework which explicitly treats collision phenomena. 
We elected to use \mujoco{}, short for ``Multi-Joint dynamics with Contact,'' for our performance baseline \cite{mujocodocs}. 
We selected \mujoco{} for the simplicity of use with Python, and the ease of reading and writing modelling parameters from its XML modelling framework. 
\mujoco{} is a simulation package which has received much attention for robot simulation \cite{erez2015simcompare,zhang2021asyncddpg,koenemann2015mpchrp2,mujoco}. 
As such, \mujoco{} has benefitted from many years of dedicated optimization in C and C++ \cite{mujocodocs},  while we implemented our integrator in Python. 
It is well-known that Python is less performant than C \cite{NumPy,prechelt-2000-CvsPy}, and it would be reasonable to not only expect \mujoco{} to outperform our algorithm in a head-on comparison of execution time, but also that our algorithm's execution time would be dominated by the runtime performance of the Python virtual machine itself. 

\begin{figure}[h]

\centering

\includegraphics[width=0.9\columnwidth]{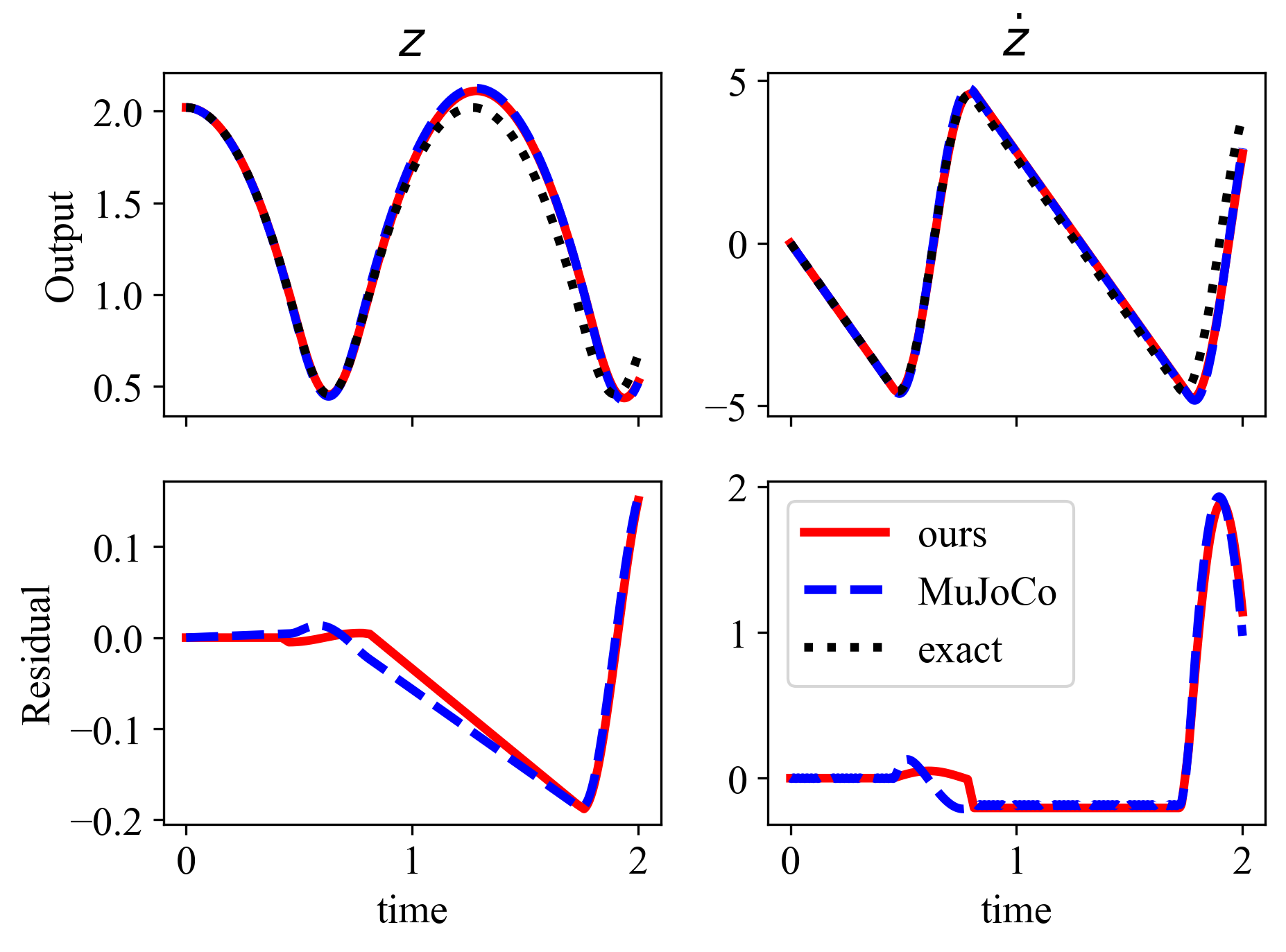}

\caption{Time series output of our calibration: integrating a the trajectory of a 1D hopping robot. 
The top row of plots show the exact trajectory (dotted gray), \mujoco{'s} result (dashed blue), and our result (red) overlaid. 
The bottom row shows the residual from the exact trajectory for both \mujoco{} and our solver. 
The left plots show height above the ground (contact height $=1$) and the right plots show vertical velocity. 
}

\label{fig:precision}

\end{figure}

To establish a meaningful comparison between our algorithm and \mujoco{} uncomplicated by the details of software implementation, we needed an event-selected toy robot model with a closed-form exact solution to calibrate both the runtime performance and the accuracy of each solver. 
Inspired by the hopping robot in Fig. \ref{fig:hopper}, we wrote a basic model for a 1D hopping robot with no damping:
\[ \dot{z} = \begin{cases} 
        -g, &z > L_0 \\
        -g + k(L_0-z)/m, &z\le L_0
    \end{cases} \]
with event functions $h_1(z)=z-z_0,\;\;h_2(z)=-z+z_0$. 
$h_2$ may seem redundant, as it has an identical guard to $h_1$, but it is easy to verify that $h_1$ only satisfies requirement (4) when $\dot{z}<0$ and that $h_2$ is needed to satisfy it when $\dot{z}>0$. 
We simulated an equivalent hopper model in \mujoco{} starting at rest with $z_0=2L_0=2$ from $t=0$ to $t=2$. 
We configured the integration step in our integrator to have the same time step ($\delta t=0.002$s) as \mujoco{}'s integrator and then ran a bisection search to find the $\varepsilon'$ where our algorithm computed a trajectory with equivalent RMS error from the exact solution as \mujoco{}, shown in Fig. \ref{fig:precision}

%
%
\begin{table}[h!]\label{tab:calibration}
    \centering
    \normalsize
    \begin{tabular}{c|c|c}
    $\varepsilon'$ & MuJoCo (s) & ours (s) \\
\hline 0.13 & 0.002436 & 0.010251
    \end{tabular}
    \caption{Calibration parameters determined by simulating the same 1D
              hopper initial value problem with \mujoco{} and our
              algorithm. 
Our algorithm achieved identical RMS integration
              error to \mujoco{} at $\varepsilon=\varepsilon'$. 
The runtime
              in seconds of each method is tabulated in the right two
              columns. 
Both solvers had maximum a time step of 0.002 seconds. 
              }
    \label{tab:runtime}
\end{table}

\subsection{Experiment}

Our algorithm's potential gains are in its treatment of multiple guards. 
We wanted to assess how its performance scaled versus \mujoco{} with increasing number of contacts. 
We developed a model of a plate in SE(2) falling due to gravity onto a bed of evenly-spaced vertical springs with frictionless rolling contacts. 
A model of this form allowed us to automatically generate models and \mujoco{} compatible XML of the plate falling onto arbitrarily many spring contacts. 
We assumed the plate was long enough that it was above each spring at all times. 
For the $i$th spring with x-position $x_{s_i}$, our flight-stance event functions and Jacobians took the form:
\begin{align*}
    h_i &= L_0 + \tan\theta\cdot(x-x_{s_i})-z \\
    Dh_i &= \begin{bmatrix} \tan\theta, & -1, & sec^2\theta \cdot(x-x_{s_i}) \end{bmatrix}
\end{align*}
for $i\in\{1,\dots,m/2\}$, with the stance-flight event functions and Jacobians being their negative as with 1D hopper. 

\begin{figure}[h]

    \centering
    
    \includegraphics[width=0.9\columnwidth]{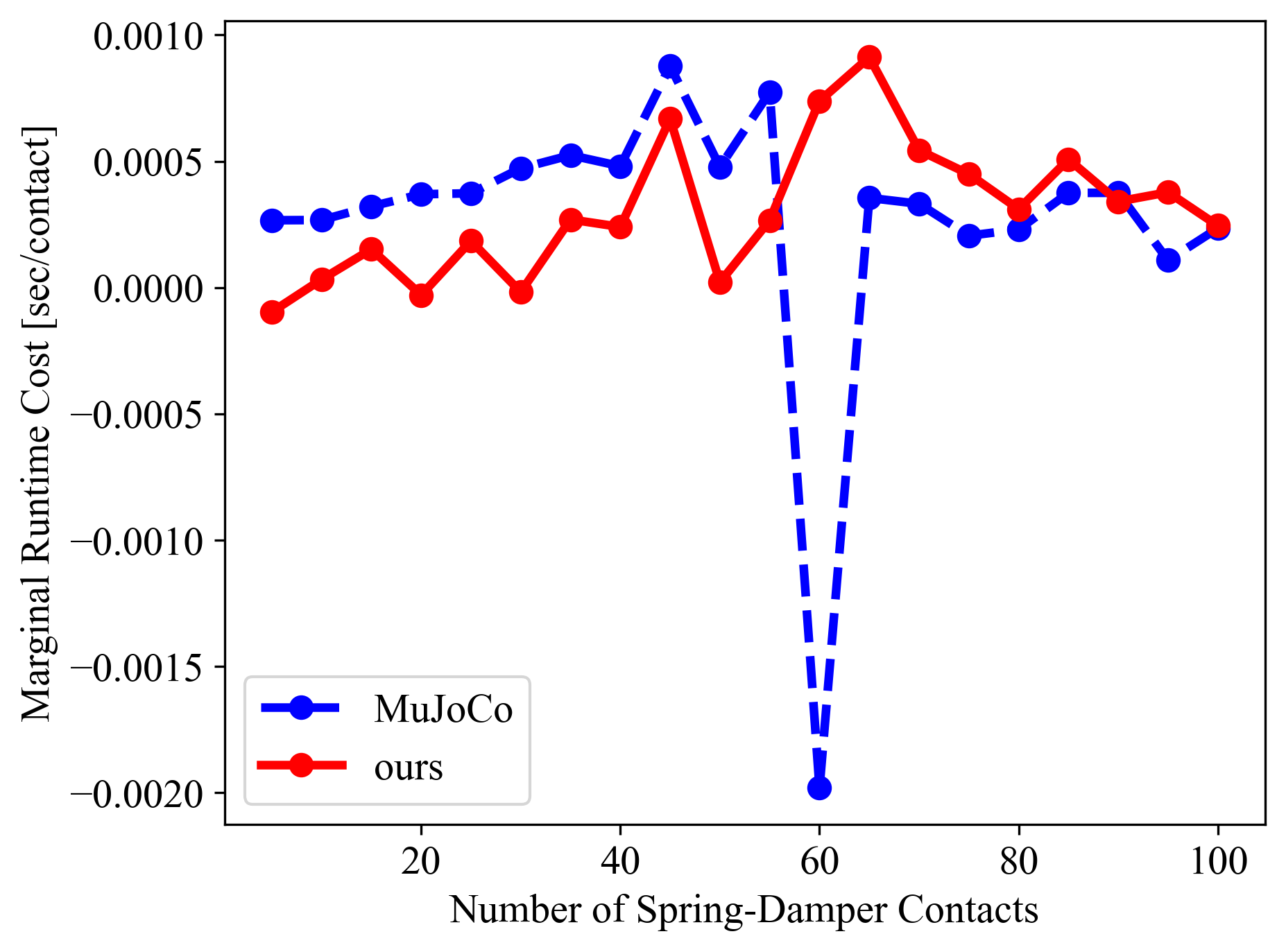}
    
\caption{The marginal cost of our algorithm (red) and \mujoco{} (dashed blue) for increasing number of spring-damper contacts. 
We integrated the same $N=120$ random initial conditions for each number of contacts. 
Our algorithm implemented in Python has roughly the same marginal cost per contact as \mujoco{}. 
}

    \label{fig:plateperf}
    
    \end{figure}

We generated 120 random initial conditions starting from rest centered over the spring bed with $z_0\in(2,3)$ and $\theta_0\in(-0.1,0.1)$. 
For each initial condition, we simulated the plate falling onto spring beds of 2 to 100 evenly-spaced identical spring-damper contacts with the precision parameter determined from our calibration. 
For a large number of contacts (roughly 150 contacts), both \mujoco{} and our algorithm started to exhibit irregular behavior, leading us to believe their performance above 100 contacts was an artifact of implementation, and not reflective of their algorithmic performance. 
We evaluated each algorithm on its marginal runtime cost per contact. 
We computed marginal cost as the derivative of the median runtime with respect to the number of contacts. 
We found that the marginal runtime cost per contact for our Python implementation was roughly the same as \mujoco{'s}, see Fig. \ref{fig:plateperf}. 

\section{Proof of First-Order Correctness}\label{sec:proof}

This ``Event Selected System'' \cite[Defn. 
2]{Burden-2015-multi} has a unique maximal flow \cite[Cor. 
5]{Burden-2015-multi} $\Phi: \Fl \to \St$ on a maximal flow domain $\Fl\subseteq \R \times \St$, and $(0,x) \in \Fl$ for all $x$. 

We denote the first parameter of $\Phi$ with a superscript, and take $\Phi^t$ to be an operator acting on a state to its right. 
Note the ``flow property'': $\Phi^t \Phi^s x = \Phi^{t+s} x$ whenever $\{ (t+s,x), (s,x), (t,\Phi^s x)\} \in \Fl$. 

Let $\tau_k:\Nei{\nu_k} \to [-\nu_k/c, \nu_k/c]$ be the unique arrival time (in future or past) $\tau_k(x)$ of initial condition $x \in \Nei{\nu_k}$ at the guard $\Gua_k$. 
This implies the existence of the impact point map $\pi_k(x) := \Phi^{\tau_k(x)} x$, $\pi_k : \Nei{\nu_k} \to \Gua_k$, which is a non-linear projection, i.e. $\pi_k \circ \pi_k = \pi_k$, $\pi_k|_{\Gua_k} = \text{Id}$. 

The correctness of our integration scheme builds on two parts: (i) a condition for the validity of projection onto the guards, which can be used for speeding up the computation of impacts on co-dimension one guard manifolds in any smooth dynamical system; and (ii) the guarantee of ESS, arising from liveness, that all transitions resolve into a unique, continuous and piecewise smooth flow \cite[Thm. 
4]{Burden-2015-multi}. 

\subsection{Efficient projection onto guards}

First, we derive a bound on the impact time for reaching all guards. 
Consider an execution starting with $x\in\Neps$, $\varepsilon\leq\min_k \nu_k$ and the arrival time $T := \tau_k(x)$ for reaching $\Gua_k$ from a starting $x$ with $h_k(x)<0$. 
Observe that
\begin{align*}
    -h_k(x) &= h_k(\Phi^T x)-h_k(x)
    = \int_0^T \frac{d}{dt} h_k(\Phi^t x) dt \\
    &= \int_0^T  \nabla h_k(\Phi^t x) \cdot f(\Phi^t x, h(\Phi^t x)) dt \geq c T, 
\end{align*}
i.e. 
the arrival time $\tau_k(x) \leq -h_k(x) / c \leq \varepsilon/c$, despite any possible domain switching induces by guard crossings. 

Consider the execution starting at $x$ with some component $h_k(x)<0$, and follow it until its earliest guard crossing, where it crosses $\Gua_j$.  
The smoothness of each of the (at most) $2^m$ pieces of $f(\cdot,\cdot)$ together the compactness of $\St$ guarantee that $f$ is Lipschitz on $\St$ with a Lipschitz constant $L$, and bounded by some bound $M$. 
Because no guards are being crossed, we can define $f_b(x) := f(x,h(x))$ the smooth vector field that applies until that first guard is crossed. 

From the commonly known \cite[Thm. 
3.4 p.96]{khalil3rdEd} we have:
\begin{align*}
    \|f_b(x) t + x - \Phi^t x\| \leq& \|x - \Phi^0 x\|\mathrm{e}^{Lt}  \\
        &+\frac{\mathrm{e}^{Lt}-1}{L}\|f_b(y)-f_b(x)\|_\infty
    \\
    \leq& \frac{2M}{L}\left(\mathrm{e}^{-h_j(x) L / c}-1\right). 
\end{align*}

Since the bound approaches zero at least linearly fast when $\varepsilon>-h_j(x)$ does, it follows that for any $\delta>0$, we select $\varepsilon>0$ such that $\|f_b(x) t + x - \Phi^t x\|<\delta$ for all $t\leq\varepsilon/c$. 
Define the first order approximate impact time $\hat \tau(x) := -h_j(x) / (\nabla h(x) \cdot f_b(x))$, and the first order approximate impact map $\hat\pi_j(x) := f_b(x)\hat\tau(x) + x$. 
We wish to bound the error $\|\hat\pi_j(x)-\pi_j(x)\|$. 

Note $\hat\tau(x) \leq -h_j(x)/c \leq \varepsilon/c$, and therefore $\|\hat\pi(x) - \Phi^{\hat\tau(x)} x\|<\delta$. 
For the true impact time we have $\|f_b(x)\tau(x) + x - \Phi^{\tau(x)} x\| = \|f_b(x)\tau(x) + x - \pi_j(x) x\|$. 
But also $\|(f_b(x)\tau(x) + x)-\hat\pi_j(x)\|=\|f_b(x)\| |\hat\tau(x)-\tau(x)|< 2 M \varepsilon/c$. 
Thus from the triangle inequality $\|\hat\pi_j(x)-\pi_j(x)\| < \delta + 2 M \varepsilon/c$, which is order $\varepsilon$ small. 

From a practical standpoint it is more convenient to give a $\varepsilon_k$ for each component $h_k(\cdot)$, such that the worst case impact point error $\|\hat\pi_k-\pi_k\|_\infty$ is sufficiently small for the simulation needed. 

The elementary calculations above provide nothing essentially novel about computing first order impact approximations compared to what is already known to the applied math community, but the approach of employing guard functions to also provide an algorithmic hint as to whether projections are appropriate to employ was not one we saw elsewhere. 
The value of our work here is that it provided a general purpose integrator that applies to ESS, which model a broad class of multi-contact problems which appear in robotics. 

\section{Discussion}
In this work we have been primarily interested in presenting our algorithm and its applicability to treating multiple contacts in robotics. 
We have provided an example implementation, a simple proof of correctness to first-order, empirically demonstrated that it can achieve second-order accuracy, and found that, with increasing number of contacts, its decrease in performance (as implemented in Python), is no worse than \mujoco{'s} decrease in performance. 
It would be interesting to see how our algorithm performs on more realistic robot models, and separately against additional simulation suites like \pybullet{} \cite{pybullet} or the Open Dynamics Engine \cite{odesim}. 
\mujoco{} is known to simulate robots faster than most other simulation suites, so we would expect our algorithm to out-perform them as well \cite{erez2015simcompare}. 
This makes our algorithm particularly well-suited to real-time applications, such as online optimization and model-predictive control of robots with 6 legs or more. 
It may now be tractable to design robot manipulators with more than 3 fingers, with real-time trajectory optimization enabled by our method. 
As state estimators are designed to be robust to noise, the speed-accuracy tradeoff of our algorithm could be more heavily exploited in contact-aware state estimation \cite{hartley2020contacIEKF,payne2020uaskf}. 

It is worth noting that ESS is more general than the types of hybrid problems \mujoco{} and \pybullet{} treat \cite{mujoco,pybullet,Burden-2015-multi}, and our algorithm could find uses outside of robotics, such as in stability analysis of power grids \cite{hiskens2000hybrid}, economics \cite{brogliato2003perspectives}, or in fields as far removed from engineering as ecology and neurobiology \cite{Burden-2015-multi,brogliato2003perspectives}.

	\bibliographystyle{siamplain}
	\bibliography{tex/my.bib}

\end{document}